\UseRawInputEncoding

\documentclass[letterpaper, 10 pt, conference]{ieeeconf}  

\IEEEoverridecommandlockouts                              

\usepackage[style=ieee,hyperref,natbib=true,backend=bibtex,firstinits,doi=false,%
     mincitenames=1,maxcitenames=2,maxbibnames=99,sorting=none,terseinits=false,hyperref=true]{biblatex}
\bibliography{references.bib}
\usepackage{amsmath,amssymb,amsfonts}
\usepackage[caption=false,font=footnotesize]{subfig}
\renewbibmacro*{bbx:savehash}{}
\defbibheading{bibliography}[\bibname]{\section*{References}}

\usepackage{hyperref}
\usepackage[capitalize]{cleveref}

\usepackage{url}
\usepackage{graphicx}
\usepackage[usenames,dvipsnames]{xcolor}
\usepackage[draft]{fixme}
\usepackage{todonotes}
\usepackage{acronym}
\usepackage{booktabs}
\usepackage{pgf} 

\usepackage{lipsum}
\usepackage{float}

\title{\LARGE \bf

Integration of the TIAGo Robot into Isaac Sim with Mecanum Drive Modeling and Learned S-Curve Velocity Profiles

}

\author{Vincent Schoenbach$^{1}$, Marvin Wiedemann$^{2}$, Raphael Memmesheimer$^{1}$, Malte Mosbach$^{1}$, and Sven Behnke$^{1}$
\thanks{$^{1}$V. Schoenbach, R. Memmesheimer, M. Mosbach, and S. Behnke are with the Autonomous Intelligent Systems group of University of Bonn, Germany; {\tt vschoenb@uni-bonn.de}}
\thanks{$^{2}$M. Wiedemann is with the Department Robotics and Cognitive Systems, Fraunhofer Institute for Material Flow and Logistics, Dortmund, Germany. }%
}

\begin{document}

\acrodef{ROS}{Robot Operating System}
\acrodef{MuJoCo}{Multi-Joint dynamics with Contact}
\acrodef{V-REP}{(Virtual Robot Experimental Platform}

\maketitle
\thispagestyle{empty}
\pagestyle{empty}


\begin{abstract}
Efficient physics simulation has significantly accelerated research progress in robotics applications such as grasping and assembly. The advent of GPU-accelerated simulation frameworks like Isaac Sim has particularly empowered learning-based methods, enabling them to tackle increasingly complex tasks. The PAL Robotics TIAGo++ Omni is a versatile mobile manipulator equipped with a mecanum-wheeled base, allowing omnidirectional movement and a wide range of task capabilities. However, until now, no model of the robot has been available in Isaac Sim. In this paper, we introduce such a model, calibrated to approximate the behavior of the real robot, with a focus on its omnidirectional drive dynamics.
We present two control models for the omnidirectional drive: a physically accurate model that replicates real-world wheel dynamics and a lightweight velocity-based model optimized for learning-based applications. With these models, we introduce a learning-based calibration approach to approximate the real robot's S-shaped velocity profile using minimal trajectory data recordings.
This simulation should allow researchers to experiment with the robot and perform efficient learning-based control in diverse environments. We provide the integration publicly at \url{https://github.com/AIS-Bonn/tiago_isaac}.
\end{abstract}


\section{Introduction} \label{sec:introduction}

General-purpose mobile manipulators hold great potential for automating a wide variety of tasks.
By combining mobility with universal manipulation capabilities, such robotic platforms can perform various tasks, including human-robot collaboration and object retrieval.
The resulting applications span industries from logistics and service robotics to healthcare.
However, realizing the full potential of mobile manipulators remains a challenge due to the complexities of integrating robust perception and control systems and the countless diverse environments such robots might face.

Physics simulation has become an essential tool in robotics research and development, as it provides a cost-effective and scalable means to train, evaluate, and refine robotic approaches before real-world deployment. 
Modern simulation pipelines allow for accelerated faster-than-real-time physics simulation, unlocking new magnitudes of data that can be generated, which is a critical component for learning-based methods on both the perception and control side. 
Additionally, simulation offers a safe environment for training and testing control algorithms, minimizing the risk of damage to both the robot and its surroundings.
These advantages have made physics simulation a cornerstone of robotic development in both academia and industry.

\begin{figure}
    \centering
    \includegraphics[width=0.9\linewidth]{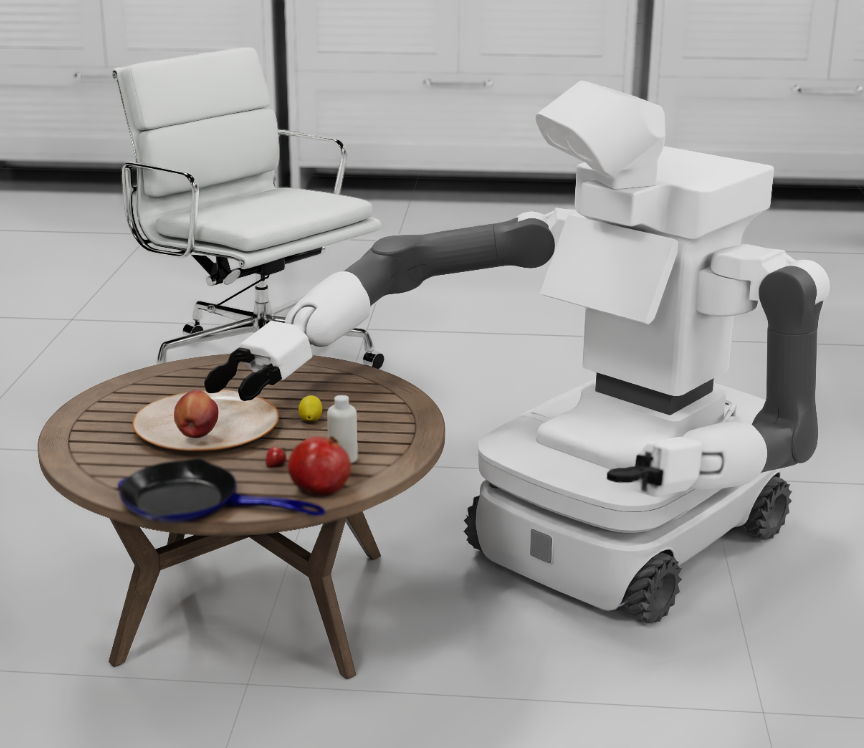}
    \caption{Sample view of the TIAGo++ Omni integrated into Isaac Sim.}
    \label{fig:sample_view}
\end{figure}

The PAL Robotics TIAGo robot \cite{pages2016tiago} is a widely adopted general-purpose mobile manipulator in the research community, valued for its versatility and close integration with the ROS framework. 
However, like many robots, TIAGo is primarily simulated in Gazebo \cite{koenig2004design}. While Gazebo has long been a standard in robotics simulation \cite{collins_review_2021}, its limited physics accuracy and graphical realism pose challenges for applications requiring precise modeling of both robot dynamics and environment perception.

Modern simulation tools, such as NVIDIA's Isaac Sim \cite{liang2018gpu}, offer significant advantages in computational efficiency, realism, and compatibility with modern machine learning pipelines. 
Its underlying physics engine leverages GPU-based parallelization—widely used in deep learning—to enable massively parallel simulations, making it well-suited for reinforcement learning (RL) and large-scale data generation. 
Furthermore, tools such as Isaac Lab \cite{mittal2023orbit}, built on Isaac Sim, provide seamless integration with RL frameworks, allowing researchers to apply state-of-the-art algorithms to their specific learning problems.

Despite these advantages, TIAGo has not yet been integrated into Isaac Sim due to several challenges. 
First, the provided Gazebo model is not directly compatible with Isaac Sim. 
Second, accurately simulating TIAGo's omnidirectional base requires precise modeling of mecanum wheels, which involves computationally expensive physics calculations. 
Third, replicating the mecanum wheel controller in simulation is difficult due to the lack of access to its internal code, making it challenging to predict how the wheels should accelerate. 
Additionally, we found that tuning PID controllers in Isaac Sim does not accurately reflect the real robot's behavior, necessitating an alternative approach.

In this work, we present an integration of the TIAGo++ Omni platform into Isaac Sim, enabling its use in modern simulation environments.
Specifically, we contribute the following:

\begin{itemize}
    \item A physically accurate model of the TIAGo++ Omni, including a high-fidelity simulation of its mecanum wheels for omnidirectional driving.
    \item As an alternative, a lightweight velocity-based control model that approximates the real robot's motion while reducing computational overhead.
    \item A neural network-based calibration approach that approximately aligns both control models with the S-shaped velocity profile of the TIAGo++ using real-world movement data.
\end{itemize}

By making our integration publicly available, we aim to facilitate research in mobile manipulation, reinforcement learning (RL), and perception.
To the best of our knowledge, no publicly available Isaac Sim model currently exists for a dual-arm omnidirectional robot.
Furthermore, we hope this integration effort serves as a guide for future robotic projects involving omnidirectional robots in Isaac Sim.

\section{Related Work} \label{sec:related_work}
Several studies have addressed the modeling of omni-wheeled platforms, though most utilize simulation tools other than Isaac Sim. 
For instance, \cite{bayar_investigation_2020} employs the Matlab-Simulink SimScape Toolbox and SolidWorks to model an omni-wheeled platform, while \cite{li_modeling_2018} uses RecurDyn for a similar purpose.
Additionally, \cite{okada_2023, apurin_omniwheel_2023} rely on Gazebo for their modeling. 
In contrast, \cite{wiedemannSimulationModelingHighly2024} leverages Isaac Sim to simulate a highly dynamic omnidirectional robot. 
Most of these studies primarily focus on kinematic modeling and validate their simulations by comparing the motion of the simulated and real robots.


Ensuring that a simulation model accurately reflects the real robot's behavior is crucial for successful sim-to-real transfer. 
Real2sim approaches \cite{hoferSim2RealRoboticsAutomation2021} optimize simulation models using real-world data.
This data is often used as a reference for manual model tuning; for example, \cite{wiedemannSimulationModelingHighly2024} enhances model accuracy by incorporating motion capture data.
Other studies go further by employing data-driven simulation models. 
Research on mathematical dynamic models, for instance, identifies model parameters using real-world data \cite{tlaleKinematicsDynamicsModelling2008a, hendzelModellingDynamicsWheeled2017}, even accounting for effects such as slipping \cite{williamsDynamicModelSlip2002}.
\cite{kanwischerMachineLearningApproach2022} applies machine learning to model an omnidirectional mobile robot, simulating its control unit, PID controllers, motors, and wheels—all trained on real-world measurements.

However, many of these studies do not account for physics engines, as common robotics simulators like Gazebo and Isaac Sim are non-differentiable. 
This non-differentiability prevents gradient-based estimation of model parameters, requiring trial-and-error strategies instead. 
For example, Evolutionary Algorithms are employed in \cite{collinsTraversingRealityGap2020} to optimize simulation models of manipulators.

To the best of our knowledge, no prior work models the S-shaped velocity profile of mecanum wheels using a neural network trained on limited, simple trajectory data recordings. 
Most existing research focuses on system identification or control using neural networks, or both, achieving strong performance in closed-loop trajectory tracking \cite{Abdalnasser2024, Ly_Thanh_Thien_Nguyen_2023}. 
However, such system identification methods are not directly compatible with physics simulators like Isaac Sim.
In contrast, our approach does not address closed-loop trajectory tracking but instead learns a lower-level velocity model—specifically, how the wheels accelerate when transitioning between motion commands. 
This allows seamless integration of our velocity model into the Isaac Sim robot simulation via the Isaac Sim API.

\section{Building the Robot Integration} \label{sec:integration}

This section describes the integration of the \textit{TIAGo++ Omni} robot into Isaac Sim, focusing on the simulation of its omnidirectional movement. 
We implement two controllers: a physically accurate controller that models full wheel dynamics and a lightweight controller that directly sets the base velocity for efficiency.
Both controllers are implemented as an Isaac Sim extension, incorporating a small neural network trained to predict the velocity curves of individual wheels.

The \textit{TIAGo++ Omni} robot is a mobile manipulator with two gripper-equipped arms, an omnidirectional mecanum wheel drive, and multiple sensors, including an RGB-D camera and LiDAR.


In the following subsections, we detail the integration of the omnidirectional drive, which posed a significant challenge due to the unique mecanum wheel mechanism.

Beyond the wheels, the remaining joints—including the torso, arms, fingers, and head—were successfully imported using the Isaac Sim URDF importer. 
Joint control is managed via ROS 2 and the Isaac Sim API, where joint state messages are sent and received.
To synchronize joint trajectory commands between the simulated and real robot, practitioners must convert trajectory commands into a sequence of joint state messages.

\begin{figure*}
  \centering
  \subfloat[Isaac Sim]{\includegraphics[height=4.8cm]{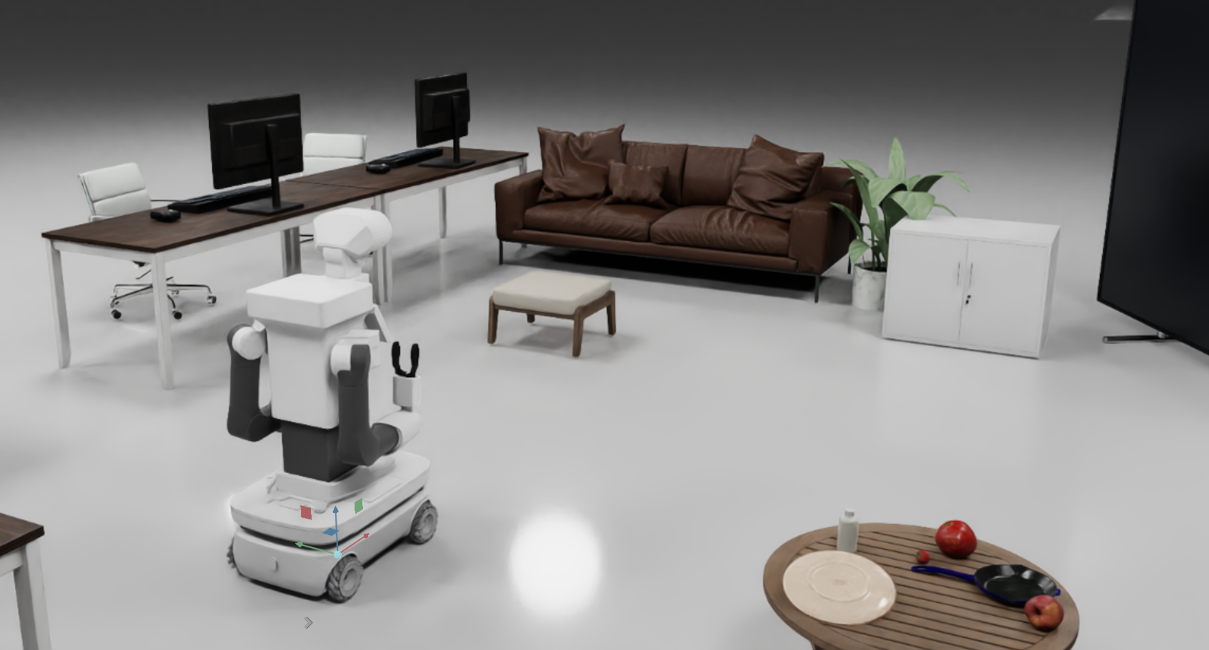}}\hfill
  \subfloat[RViz Visualization]{\includegraphics[height=4.8cm]{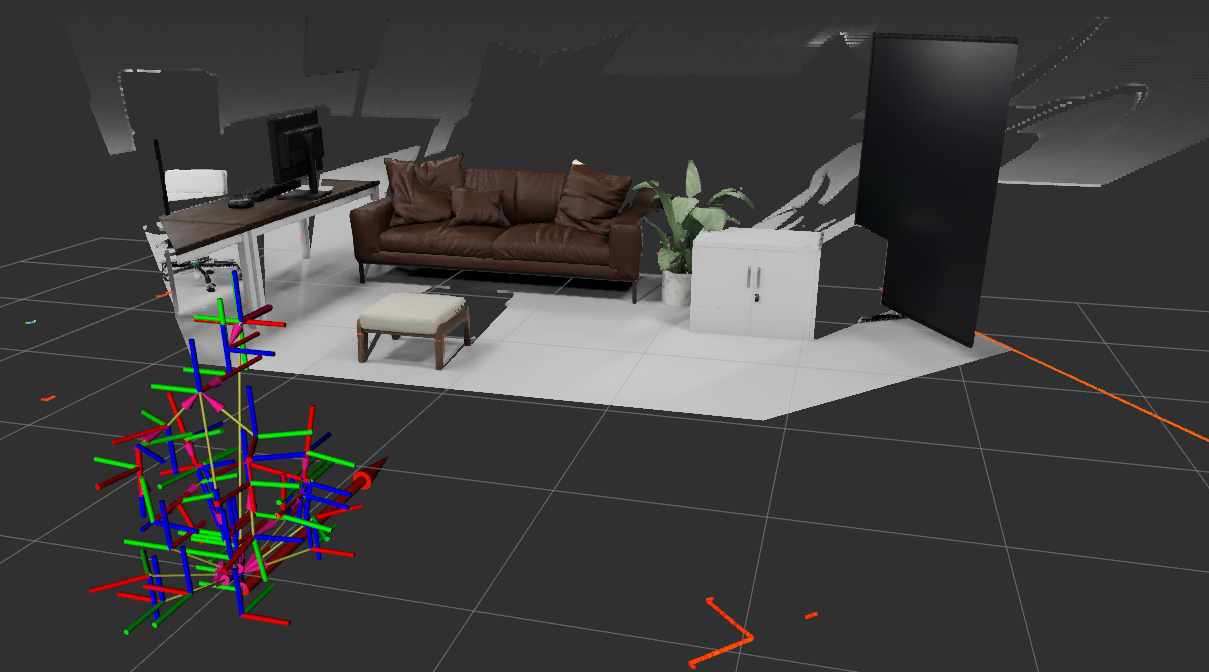}}
  \caption{(a) Simulation in Isaac Sim and (b) corresponding sensor data visualization in RViz 2.}
  \label{fig:comparison_with_rviz}
\end{figure*}

Integrating sensors such as LiDAR and cameras into Isaac Sim is straightforward. 
The simulator provides various built-in sensors, and using the Isaac Sim API, sensor data can be recorded and published to ROS 2 topics. 
Example sensor data and the corresponding transform tree visualization in RViz 2 are shown in \cref{fig:comparison_with_rviz}.

\subsection{Integration of the Omnidirectional Drive by Full Physical Simulation}

In the Gazebo simulation provided by PAL Robotics, the TIAGo robot's omnidirectional drive is approximated rather than physically simulated \cite{pal_gazebo_tiago_simulation}. While some efforts in Gazebo aim to improve omnidirectional drive modeling \cite{apurin_omniwheel_2023}, mecanum and similar wheels are typically not simulated with full physical accuracy. As a result, direct omnidirectional driving in Isaac Sim is not feasible using the imported \texttt{.urdf} file and requires a custom extension.

We address this by implementing a physically accurate simulation based on \cite{wiedemannSimulationModelingHighly2024}, following three key steps: (i) wheel modeling, (ii) roller collider modeling, and (iii) implementation of the holonomic controller in Isaac Sim.

\begin{figure}[H]
    \centering
    \subfloat[Procedurally generated mecanum wheel.]{%
        \includegraphics[height=2.05cm]{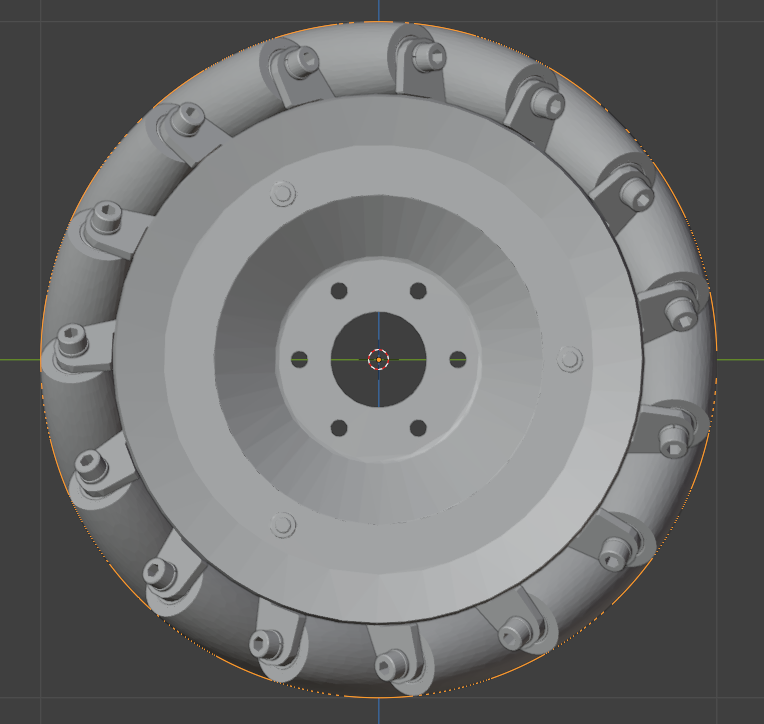}
        \label{fig:wheels}
    }
    \hfill
    \subfloat[Roller collider modeled with six spheres, balancing computational efficiency and physical accuracy.]{%
        \includegraphics[height=2.05cm]{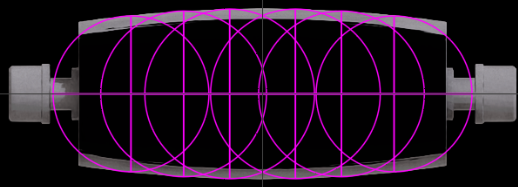}
        \label{fig:roller_collider}
    }
    \caption{Model of (a) the mecanum wheel and (b) the roller colliders.}
    \label{fig:combined}
\end{figure}

For mecanum wheel modeling, we replaced the wheels from the \texttt{.urdf} file with custom ones (\cref{fig:wheels}) generated using the script from \cite{fuji_mecanum_wheels}. These wheels feature 15 free-spinning rollers angled at 45 degrees and are scaled to match the original size. Proper roller alignment and a precisely round wheel shape are crucial, as misalignment can lead to loss of ground contact and unintended jumps in driving behavior \cite{dickerson_control_1991}. The generated wheels in our simulation ensure these requirements are met.

A crucial aspect of the simulation is the roller collider design, as ground contact directly affects the robot's motion, particularly for lateral driving. Following the methodology in \cite{wiedemannSimulationModelingHighly2024}, we modeled each roller collider using six spheres (\cref{fig:roller_collider}). 

This approach ensures smooth motion in Isaac Sim while being significantly more computationally efficient than mesh-based or low-poly collision models \cite{apurin_omniwheel_2023}.

To control the robot omnidirectional, the wheel velocities for all four wheels must be computed from the movement command, which is provided as a \texttt{Twist} message. A \texttt{Twist} command specifies linear velocities $v_x, v_y$ along the x- and y-axes, as well as rotational velocity $v_\theta$ around the robot's center. 

If $r$ denotes the wheel radius, and $L_x, L_y$ represent the distances from the robot's center to the wheels along the x- and y-axes, then the relationship between the base velocity $(v_x, v_y, v_\theta)$ and the four angular wheel velocities $\omega_i$ can be derived to be (see e.g., \cite{KinematicModel2015}):

\begin{equation} \label{matrixformula}
\begin{bmatrix} \omega_1 \\ \omega_2 \\ \omega_3 \\ \omega_4 \end{bmatrix} =
\frac{1}{r}
\begin{bmatrix} 
1 & 1 & -(L_x + L_y) \\ 
1 & -1 & (L_x + L_y) \\ 
1 & -1 & -(L_x + L_y) \\ 
1 & 1 & (L_x + L_y) 
\end{bmatrix}
\begin{bmatrix} v_x \\ v_y \\ v_\theta \end{bmatrix}.
\end{equation}

While this formulation allows the robot to move in the desired direction, it does not fully replicate the real robot's base movement, which behaves as a black box—i.e., the exact mapping from velocity commands to wheel accelerations is unknown. Notably, the real robot's controller exhibits S-shaped velocity curves (\cref{fig:s_shape_velocity_curve_fit}), but since we lack access to the internal controller, we cannot directly inspect its behavior. 

As the matrix formulation alone does not account for these acceleration dynamics, we recorded real-world trajectory data and trained a small neural network to predict the required acceleration profiles for given movement commands, as described in the following section.

\subsection{Predicting the Wheel Velocity Profile from Data} \label{sec:predicting_the_wheel_velocity_profile_from_data}
\begin{figure}[H]
    \centering
    \resizebox{1.\linewidth}{!}{\input{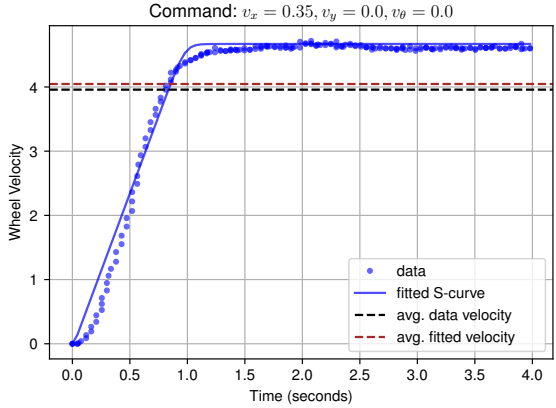}}
    \caption{Example of an S-shaped velocity curve fitted to wheel velocity data. The velocity command is to move in the x-direction at $0.35$ m/s. Since all wheels behave identically in this scenario, only one wheel's velocity is shown.}
    \label{fig:s_shape_velocity_curve_fit}
\end{figure}

To approximate the smooth acceleration profile of the real robot for any given target wheel velocity, we train a neural network to predict an S-shaped velocity curve that closely matches the recorded data (\cref{fig:s_shape_velocity_curve_fit}).

A detailed explanation of the mathematical model that best matches the recorded data, identified through trial and error, is provided in Appendix~\ref{sec:appendix_wheel_velocity_profile_model}. Setting aside the intricacies, we predict the parameters $\Theta$ of an S-shaped function $S_{\Theta}:\mathbb{R}_{\geq 0} \rightarrow \mathbb{R}_{\geq 0}$ based on a given target wheel velocity $\omega$. This function maps elapsed time to velocity according to the learned S-shaped velocity profile. 

To achieve this, we train a small neural network with weights $W$ to approximate the mapping $\Theta = f_W(\omega)$ using data from transitions starting at zero velocity. At runtime, we apply the model to arbitrary transitions by feeding in the wheel velocity difference $\Delta \omega$ between the current and target states. The network then outputs the corresponding S-curve parameters $\Theta$, allowing us to smoothly interpolate between any two commands. Because different velocity changes result in distinct acceleration profiles, the goal is for the network to generalize from the training data and predict suitable S-curve parameters for any given target velocity $\omega$, or more generally, any velocity change $\Delta \omega$.

The training data consists of recorded wheel velocity profiles from movements in the x-direction, y-direction, and rotational motion around the z-axis, ensuring coverage of the robot's basic degrees of freedom. Each command generates a unique velocity profile for the four wheels. For details on the recorded trajectories, see Section~\ref{sec:experiments}.

We observe that collecting individual wheel acceleration profiles for various target velocities results in a relatively consistent family of S-curves (\cref{fig:s_shape_velocity_curve_all_data}(a)). Notably, even for significantly different commands—such as x-direction movement versus rotational movement—some wheel velocity profiles align on up to two wheels due to the kinematics of mecanum-wheeled driving.

\begin{figure*}
  \centering
  \subfloat[Selection of recorded data]{\resizebox{0.45\linewidth}{!}{\input{images/s_curve_data.pgf}}}\quad  
  \subfloat[Predicted S-curves for various target wheel velocities]{\resizebox{0.45\linewidth}{!}{\input{images/s_curves_predicted.pgf}}}\hfill
  \caption{(a) Wheel velocity curves for different commands in the x-direction (red curves), y-direction (blue curves), and rotational motion (green curves). The S-curve shape exhibits some inconsistencies, likely due to noise from the real robot's PID controller. (b) Predicted S-curves from the fitted model.}
  \label{fig:s_shape_velocity_curve_all_data}
\end{figure*}

Ideally, on a mecanum-wheeled robot, the velocity profile for a given \texttt{Twist} command should ensure proportional acceleration across all wheels. That is, each wheel should follow an acceleration curve scaled to its final target velocity, preserving the correct velocity ratios throughout the transition. This ensures that the robot moves in the intended direction not only once all wheels reach their target speeds, but also during the acceleration phase itself. Deviations from proportional acceleration can result in unintended drift or curvature during motion. This observation suggests that the acceleration behavior of individual wheels should be effectively modeled based on their target velocity change, independent of the full \texttt{Twist} command—a principle we follow in our approach above.

\begin{figure}[H]
    \centering
    \resizebox{1.\linewidth}{!}{\input{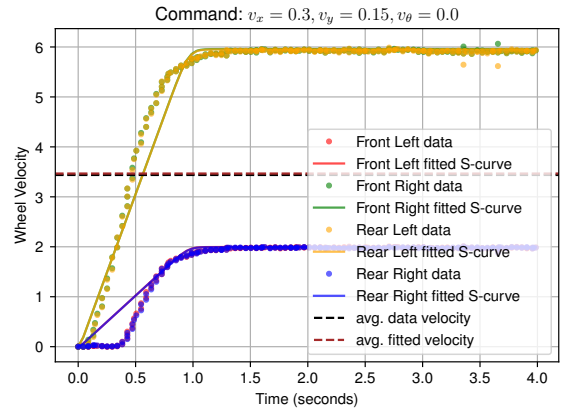}}
    \caption{For non-axis-aligned movements, such as a skewed trajectory in the direction $x = 2y$, the predicted acceleration behavior of the four wheels deviates from the real robot due to the original controller not enforcing proportional acceleration. However, the average velocity of our learned model remains close to the expected value.}
    \label{fig:s_curve_xy_prediction_failure}
\end{figure}

However, in practice, we found that PAL Robotics, the manufacturer of the TIAGo robot, did not fully adhere to this principle, resulting in non-proportional velocity profiles for certain commands, such as movements combining x- and y-directional motion (\cref{fig:s_curve_xy_prediction_failure}). Consequently, our S-curve predictions for such movements exhibit deviations due to the suboptimal design of the original robot controller. Additionally, we observed that the real robot exhibits slight deviations from the intended direction during acceleration.

The most effective solution would be to reimplement the original robot controller to enforce proportional acceleration. However, since we do not have access to this level of hardware control, this task would fall to PAL Robotics. More broadly, this highlights an important consideration when designing controllers for mecanum-wheeled robots.

Our strategy for simulating a movement command is as follows: Suppose the robot is currently at velocity $T = (v_x, v_y, v_\theta)$ and the desired velocity is $T' = (v_x', v_y', v_\theta')$. To ensure a smooth transition, we interpolate the commands received by the simulated robot between $T$ and $T'$.

To do this, first, we compute the total required velocity change $\Delta \omega$ for the wheels (see Appendix~\ref{sec:appendix_TotalVelocityChangeComputation} for details) and use this as input to our model to obtain the velocity S-curve, parameterized by $\Theta = f(\Delta \omega)$. This curve is then used to update the target \texttt{Twist} command at each timestep:

\begin{equation}
    T_t = T + \text{p}_t \cdot (T' - T),
\end{equation}

where 

\begin{equation}
    \text{p}_t = \frac{S_{\Theta}(t - t_0)}{\Delta \omega}
\end{equation}

represents the fraction of the command executed at time $t$, with $t_0$ denoting the command start time.

The wheel controllers then receive target velocities derived from the updated \texttt{Twist} command $T_t$, which is converted using Equation~\ref{matrixformula}.

\begin{figure*}
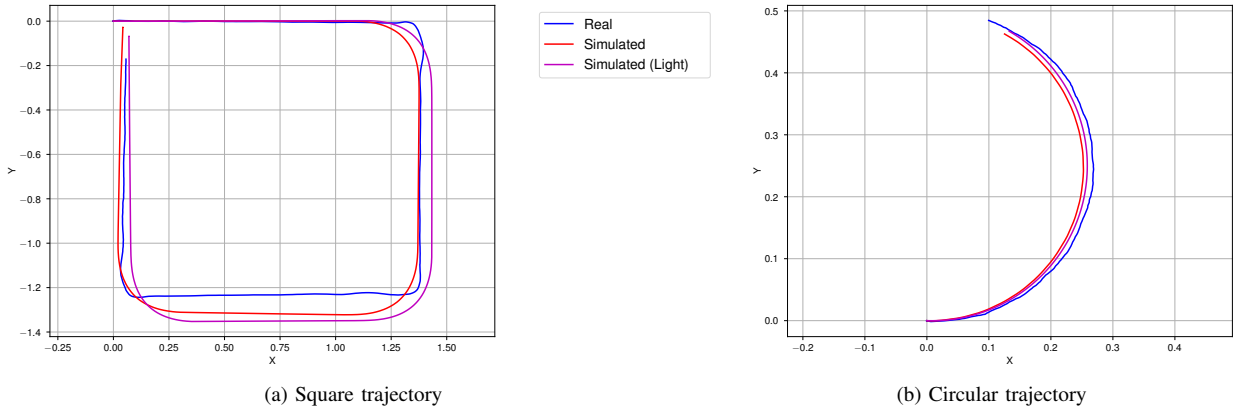

  \centering
  \subfloat[Square trajectory]{
  \resizebox{!}{5cm}{\input{images/square_trajectory.pgf}}}\quad
  \subfloat[Circular trajectory]{\resizebox{!}{5cm}{\input{images/19635.pgf}}}\hfill
  \caption{Comparison of simulated and real robot trajectories: (a) Square trajectory consisting of 3-second movements in forward, rightward, backward, and leftward directions with a target velocity of $0.45$ m/s, and (b) Circular trajectory with $0.19$ m/s movement in the y-direction and $0.78$ rad/s rotational movement.\newline Note that our experiments do not use closed-loop control, meaning the robot has no feedback on its actual position during execution.}
  \label{fig:experiment_example_trajectories}
\end{figure*}

Overall, this heuristic approach for interpolating between the current velocity $T$ and the desired velocity $T'$ ensures a smooth transition, replicating the S-shaped acceleration profiles observed in the real robot's basic movements. However, while our approach provides a realistically looking controller for simulation, the real robot does not exhibit identical transitional behavior and may even accelerate more rapidly. For example, the corners of a square trajectory—where the robot undergoes a 90° change in movement direction—are navigated much more smoothly in simulation than on the real robot (\cref{fig:experiment_example_trajectories}(a)). Consequently, the simulated robot is slightly harder to control than the real robot.

Nevertheless, we argue that this discrepancy is not a major concern for two reasons. First, making the simulation more challenging to control than the real robot introduces artificial noise, which can improve robustness. Second, since our controller operates in an open-loop fashion, a higher-level closed-loop trajectory controller should still be able to utilize our open-loop movement controller effectively.

We also hypothesize that accurately modeling the acceleration profile of an arbitrary black-box mecanum robot controller may be nearly impossible, or at least highly challenging, as it would require collecting extensive real-world data for transitions between all possible velocity states and fitting a model to predict the resulting acceleration behavior for each wheel. We leave this question open for future research but, as we argued, believe that our approach is already sufficient for most use cases. However, future work using our model would have to confirm whether this is true.

\subsection{Integration of the Omnidirectional Drive by Directly Setting the Base Velocity}

As an alternative to the physically accurate simulation of mecanum wheels, we provide a lightweight approach to simulating omnidirectional movement. Instead of modeling wheel-ground interactions, this method directly sets the robot's base velocity to the desired \texttt{Twist} command using the Isaac Sim API. While we still apply the S-shaped velocity profile for smooth acceleration, individual wheel velocities are no longer explicitly controlled.

This method is less physically accurate, as it neglects forces such as friction and does not even require knowledge of how mecanum wheels function. However, as demonstrated in the Experiments section, it remains sufficiently accurate while significantly improving simulation efficiency. By eliminating roller collision computations and treating the wheels as dummy components, we achieve a notable performance gain, e.g., reducing the required physics steps per second from 360 to 60. In practice, this leads to a clear reduction in computational cost, which is noticeably reflected in higher simulation frame rates (FPS).

A key advantage of this approach is its generalizability --- it can be applied to any holonomic robot, regardless of the real PID controller's implementation. Moreover, it eliminates the need to tune wheel joint physics parameters in Isaac Sim. Additionally, if only the lightweight controller is used, the mecanum wheels do not need to be modeled at all.

\section{Experiments} \label{sec:experiments}

In this section, we compare the behavior of the simulated robot with that of the real robot. To do so, we recorded multiple simple open-loop trajectories in both environments and analyzed their differences. Specifically, we compared the simulation's odometry data to motion tracking data from the real robot executing the same trajectories. Motion tracking was conducted using OpenVR with a VIVE tracker mounted at the robot's rotational center, while the robot's pose (position and orientation) was triangulated using two VIVE lighthouses.

To evaluate simulation accuracy, we recorded various simple trajectories for both simulated versions and the real robot. 

Among our experiments, we tested 20 different target velocities ranging from $0.05$ to $1.00$ m/s in the x-, y-, and x-y- (diagonal) directions, as well as target rotational velocities around the z-axis from $0.05$ to $1.5$ rad/s. Each command lasted four seconds, during which we recorded the resulting wheel velocity profiles using the robot's onboard encoders. To ensure smooth and consistent data, each command was repeated three times, and the resulting profiles were averaged. The relative errors for each case are summarized in Table~\ref{tab:error_summary}. Overall, these simulated simple trajectories closely follow the intended paths of the real robot's controller.

\begin{table}[H]
    \centering
    \caption{Mean relative error (MRE) and standard deviation (STDRE) for 4-second velocity commands}
    \begin{tabular}{lcc}
        \toprule
        \textbf{MRE $\pm$ STDRE} & \textbf{\% Value} & \textbf{\% Value (Light)} \\
        \midrule
        x-direction & $8.24 \pm 1.37$ & $7.36 \pm 3.71$ \\
        y-direction & $4.61 \pm 5.54$ & $3.89 \pm 1.49$ \\
        x-y-direction & $5.68 \pm 2.71$ & $5.16 \pm 3.07$ \\
        Rotation & $4.30 \pm 1.62$ & $2.97 \pm 1.52$ \\
        \bottomrule
    \end{tabular}
    \label{tab:error_summary}
\end{table}

The table shows mean relative error (MRE) and standard deviation (STDRE) for 4-second velocity commands in the x-, y-, and x-y-directions, as well as for rotation. The relative error is defined as $\frac{| \Delta p_{\text{real}} - \Delta p_{\text{sim}} |}{\Delta p_{\text{real}}}$, where $\Delta p_{\text{real}}$ and $\Delta p_{\text{sim}}$ denote the total traveled distances for linear motion (or total rotation angles for rotational motion), measured on the real and simulated robots, respectively. The MRE and STDRE represent the mean and standard deviation of this error across all tested velocities.

Note that diagonal (x-y) commands were used for evaluation only and were not included in the training data, allowing us to assess generalization to combined motions.

Figure~\ref{fig:experiment_example_trajectories} presents two example trajectories. In the first, the robot moves forward, rightward, backward, and leftward, forming an approximate square. In the second, the robot moves forward while rotating leftward, tracing a circular path.

For longer and more complex trajectories, accumulated errors become more significant, particularly when certain acceleration patterns, such as simultaneous x-y directional movement, are not modeled as accurately (as we saw before in \cref{fig:s_curve_xy_prediction_failure}). However, despite these limitations, the S-curve approximation remains sufficiently accurate for pure movements, as indicated by the relatively low error for x-y-directional motion in Table~\ref{tab:error_summary}. 

In practice, the slight error accumulation in the open-loop trajectories is unlikely to be a major issue, as real-time closed-loop trajectory controllers continuously correct errors and adjust movement commands to maintain the intended path.

\section{Conclusion}

In this work, we presented the integration of the TIAGo++ Omni robot into Isaac Sim, enabling its use in modern simulation environments. While the primary focus was the modeling of mecanum wheel dynamics, we also implemented a full robot simulation, including sensor and joint control integration with ROS 2, ensuring compatibility with existing frameworks.  

For the omnidirectional drive, we introduced two simulation models: a physically accurate model that closely replicates real-world wheel dynamics and a lightweight velocity-based model that significantly improves simulation efficiency. The lightweight model is well-suited for high-throughput tasks such as reinforcement learning, while the physically accurate model can be used for fine-tuning and validation when precise behavior is required. Additionally, our learning-based calibration improves simulation accuracy by approximating the S-shaped velocity profile of the real robot using straightforward data recordings.  

A key insight from our work is that ensuring proportional acceleration across all wheels is crucial for smooth omnidirectional movement in mecanum-wheeled robots. Our findings suggest that non-proportional acceleration patterns can lead to inconsistencies and driving behavior that is difficult to model and replicate in simulation, highlighting an important design consideration for controllers of mecanum-wheeled bases.

Future work may focus on further validating the simulation framework for tasks such as reinforcement learning and trajectory tracking. Additionally, alternative approaches could be explored to improve the modeling of S-shaped velocity profiles for mecanum-wheeled robots.

\section*{Acknowledgment}
This research has been partially funded by the Federal Ministry of Education and Research of Germany under grants no. 16ME0999 RIG and 16SV8683 RimA.
The authors acknowledge the use of OpenAI's ChatGPT-4o for language refinement, including grammar, clarity, and conciseness improvements. All technical content, results, and interpretations remain the sole work of the authors.

\printbibliography
\newpage

\appendix

\subsection{Model of the Learned S-Curves} \label{sec:appendix_wheel_velocity_profile_model}

Here, we describe the selection of our S-curve model $S_\Theta$, used in Section~\ref{sec:predicting_the_wheel_velocity_profile_from_data}. Our goal was to choose a function that best aligns with the inductive bias observed in the data, i.e., an S-curve that closely matches the velocity profiles of the robot.

In general, an S-curve can be described by a set of parameters $\Theta \in \mathbb{R}^d$. A common example is the logistic function, parameterized by its growth rate and midpoint. A small neural network, e.g., a multi-layer perceptron (MLP), $f:\mathbb{R} \rightarrow \mathbb{R}^d$, learns to predict the parameters $\Theta_\omega$ for each target wheel velocity $\omega$. As long as the S-curve is differentiable, we can backpropagate through it, effectively treating it as the final layer of the network.

However, after experimenting with various known S-curves, such as the logistic function and others, we found that none captured the observed behavior as effectively as the custom-assembled S-curve we describe next. The reason is that the velocity profiles exhibit asymmetrical sharpness, with a steeper initial rise, a smoother asymptotic approach to the final velocity, and a linear transition in between. 

\begin{figure}
    \centering
    \resizebox{1.\linewidth}{!}{\input{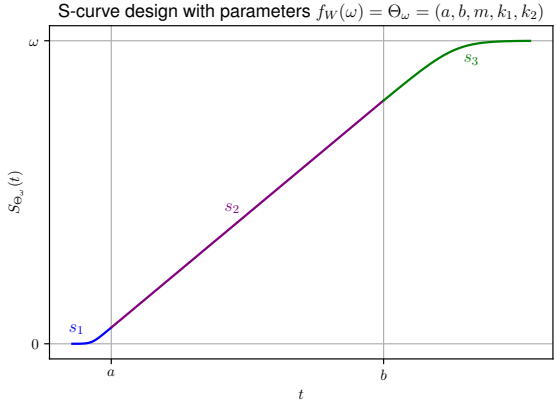}}
    \caption{Structure of the learned S-curve model $S_{\Theta_\omega}$ for an example target velocity $\omega=11.48$, using the actual learned weights $W$.}
    \label{fig:S_curve_model}
\end{figure}

Our S-curve model $S_{\Theta_\omega}$ (Figure~\ref{fig:S_curve_model}) is parameterized by five variables, $\Theta_\omega = (a, b, m, k_1, k_2)$, and consists of three segments:

\begin{enumerate}
    \item Initial phase: A smooth-ramp-like function ${s_1}_{a, k_1}$ in the interval $[0, a]$.
    \item Linear transition: A linear function ${s_2}_{a, b, m}$ in the interval $[a, b]$.
    \item Final phase: A flipped smooth-ramp-like function ${s_3}_{b, k_2}$ in the interval $[b, \infty)$.
\end{enumerate}

The parameters $a$ and $b$ determine the transition points between these segments, $k_1$ and $k_2$ control the sharpness of the functions $s_1$ and $s_3$, and $m$ defines the slope of the linear segment $s_2$.

The ramp-like functions $s_1$ and $s_3$ are implemented as softplus functions of the form:
\begin{equation}
    s(x) = \frac{\log(1 + \exp(x \cdot \text{sharpness}))}{\text{sharpness}}
\end{equation}
However, any other ramp-like functions could also have been used.

During training, we enforce continuity by ensuring that:
\begin{enumerate}
    \item $s_1$ and $s_2$ meet at $a$.
    \item $s_2$ and $s_3$ meet at $b$.
    \item $s_1$ and $s_3$ asymptotically approach the same slope $m$ as $s_2$.
    \item $s_1(0) = 0$ and $s_3(\infty) = \omega$.
\end{enumerate}

The implementation details of enforcing continuity constraints are available through our open-source code, though alternative implementations could achieve similar results.

We designed our neural network architecture as a small MLP with layers sized $[1,35,15,5]$, employing softplus activation functions between layers. The network predicts the five parameters defining the S-curve for a given target wheel velocity. This compact network structure was chosen because it provides a good balance between fitting accuracy and inference efficiency.

We chose the Adam optimizer arbitrarily for training. To maximize performance, we trained 100 network instances with different random initializations and selected the best-performing model based on the lowest validation error. We do not claim that our model architecture or training procedure is optimal; rather, we just want to highlight that our chosen approach achieves good performance.


\subsection{Computation of Total Velocity Change Bound}
\label{sec:appendix_TotalVelocityChangeComputation}

The quantity $\Delta v_{\mathrm{bound}}$ represents a conservative upper bound on the velocity change at any of the wheel contact points due to both translational and rotational motion. Since the mecanum wheels are positioned at fixed offsets $L_x$ and $L_y$ from the robot’s center, any change in the chassis velocities affects the wheel speeds differently depending on their location. By the triangle inequality, each signed component of the true per‐wheel velocity change is bounded in magnitude by the sum of the magnitudes of its translational and rotational contributions.

Let $v_x, v_y, v_\theta$ be the current chassis velocities in the $x$‐, $y$‐, and rotational directions, respectively, and let $v_x', v_y', v_\theta'$ be the commanded velocities.  We define
\begin{align}
\Delta v_{x,\mathrm{bound}}
&= \bigl|\,v_x' - v_x\bigr| \;+\;\bigl|\,v_\theta' - v_\theta\bigr|\;L_y,\\
\Delta v_{y,\mathrm{bound}}
&= \bigl|\,v_y' - v_y\bigr| \;+\;\bigl|\,v_\theta' - v_\theta\bigr|\;L_x.
\end{align}
Then the combined chassis‐wide bound is
\begin{equation}
\Delta v_{\mathrm{bound}}
= \sqrt{\Delta v_{x,\mathrm{bound}}^{2}
       +\,
       \Delta v_{y,\mathrm{bound}}^{2}}
\end{equation}
and normalizing by the wheel radius $r$ gives the angular‐speed bound
\begin{equation}
\Delta\omega
:= \frac{\Delta v_{\mathrm{bound}}}{r}.
\end{equation}

In our approach, we then use $\Delta\omega$ as the single feed‐forward input to our S‐curve model.

\end{document}